\documentclass{article}

\usepackage[preprint]{preprint}

\usepackage[utf8]{inputenc} %
\usepackage[T1]{fontenc}    %
\usepackage{hyperref}       %
\usepackage{url}            %
\usepackage{booktabs}       %
\usepackage{amsfonts}       %
\usepackage{amsmath,amssymb,amsthm}
\usepackage{nicefrac}       %
\usepackage[dvipsnames,table]{xcolor} %
\usepackage{graphicx}
\usepackage{subcaption}
\usepackage{tikz}
\usetikzlibrary{positioning,arrows.meta}

\usepackage{booktabs,array,tabularx}
\usepackage{enumitem}

\ifdefined\DeclareUnicodeCharacter
\DeclareUnicodeCharacter{00B7}{\ensuremath{\cdot}}       %
\DeclareUnicodeCharacter{00D7}{\ensuremath{\times}}      %
\DeclareUnicodeCharacter{2190}{\ensuremath{\leftarrow}}  %
\DeclareUnicodeCharacter{2192}{\ensuremath{\rightarrow}} %
\DeclareUnicodeCharacter{2264}{\ensuremath{\leq}}        %
\DeclareUnicodeCharacter{2265}{\ensuremath{\geq}}        %
\DeclareUnicodeCharacter{2260}{\ensuremath{\neq}}        %
\DeclareUnicodeCharacter{2200}{\ensuremath{\forall}}     %
\DeclareUnicodeCharacter{2014}{\textemdash}              %
\DeclareUnicodeCharacter{27E8}{\ensuremath{\langle}}     %
\DeclareUnicodeCharacter{27E9}{\ensuremath{\rangle}}     %
\DeclareUnicodeCharacter{2227}{\ensuremath{\wedge}}      %
\DeclareUnicodeCharacter{00AC}{\ensuremath{\neg}}        %
\DeclareUnicodeCharacter{2203}{\ensuremath{\exists}}     %
\DeclareUnicodeCharacter{2208}{\ensuremath{\in}}         %
\DeclareUnicodeCharacter{22A2}{\ensuremath{\vdash}}      %
\DeclareUnicodeCharacter{2194}{\ensuremath{\leftrightarrow}} %
\DeclareUnicodeCharacter{21A6}{\ensuremath{\mapsto}}     %
\DeclareUnicodeCharacter{2115}{\ensuremath{\mathbb{N}}}  %
\DeclareUnicodeCharacter{2124}{\ensuremath{\mathbb{Z}}}  %
\DeclareUnicodeCharacter{03B1}{\ensuremath{\alpha}}      %
\DeclareUnicodeCharacter{03B2}{\ensuremath{\beta}}       %
\DeclareUnicodeCharacter{03C3}{\ensuremath{\sigma}}      %
\DeclareUnicodeCharacter{03A3}{\ensuremath{\Sigma}}      %
\DeclareUnicodeCharacter{2026}{\ensuremath{\ldots}}      %
\DeclareUnicodeCharacter{2212}{\ensuremath{-}}           %
\DeclareUnicodeCharacter{2080}{\ensuremath{{}_{0}}}      %
\DeclareUnicodeCharacter{2081}{\ensuremath{{}_{1}}}      %
\DeclareUnicodeCharacter{2082}{\ensuremath{{}_{2}}}      %
\DeclareUnicodeCharacter{2098}{\ensuremath{{}_{m}}}      %
\DeclareUnicodeCharacter{25B8}{\ensuremath{\blacktriangleright}} %
\DeclareUnicodeCharacter{0101}{\ensuremath{\bar{a}}}     %
\DeclareUnicodeCharacter{2983}{\ensuremath{\{\!|}}       %
\DeclareUnicodeCharacter{2984}{\ensuremath{|\!\}}}       %
\fi

\newcommand{\li}[1]{\texttt{#1}}        %

\theoremstyle{plain}

\theoremstyle{definition}

\usepackage{mathtools}
\usepackage{wrapfig}
\usepackage{placeins}

\hypersetup{
  pdftitle={Eureka: Lean Proofs as a Public Research Service},
  pdfauthor={Debargha Ganguly et al.},
  pdfkeywords={Lean, formal verification, theorem proving, open-weight models, agents, PutnamBench},
  colorlinks=true,
  linkcolor=BrickRed,
  citecolor=MidnightBlue,
  urlcolor=MidnightBlue,
}
\usepackage[capitalise, noabbrev, nameinlink]{cleveref}

\definecolor{oursrow}{HTML}{EAF1FB}            %
\newcommand{\ours}{\rowcolor{oursrow}}
\newcommand{\best}[1]{\textbf{#1}}              %
\newcommand{\na}{\textcolor{black!30}{--}}      %
\newcommand{\budget}[1]{\,{\scriptsize\textcolor{black!55}{#1}}}  %
\definecolor{goodgreen}{HTML}{1A7F37}
\definecolor{badred}{HTML}{B42318}
\definecolor{mixamber}{HTML}{B54708}
\definecolor{heat}{HTML}{3BAA6A}
\newcommand{\gain}[1]{\,{\scriptsize\textcolor{goodgreen}{$\blacktriangle$#1}}}  %
\newcommand{\xours}[1]{\,{\scriptsize\textcolor{badred}{#1$\times$}}}             %
\newcommand{\heat}[2]{\cellcolor{heat!#1}#2}                                    %
\newcommand{\wopen}{\textcolor{goodgreen}{open}}
\newcommand{\wclosed}{\textcolor{black!55}{closed}}
\newcommand{\wmixed}{\textcolor{mixamber}{mixed}}
\newcommand{\wnd}{\textcolor{black!35}{n/d}}
\definecolor{catlean}{HTML}{E69F00}
\definecolor{catshell}{HTML}{009E73}
\definecolor{catfile}{HTML}{CC79A7}
\definecolor{catretr}{HTML}{56B4E9}
\definecolor{catagent}{HTML}{0072B2}
\newcommand{\tshare}[2]{\makebox[22pt][r]{#2\%}\enspace\textcolor{#1}{\rule[-0.1ex]{\fpeval{0.5*#2}pt}{1.3ex}}}
\newcommand{\tsmall}{\makebox[22pt][r]{${<}0.1\%$}}
\newcommand{\tcat}[2]{\rowcolor{#1!12}\multicolumn{5}{l}{\textcolor{#1}{$\blacksquare$}\,\textbf{#2}}\\}

\newcommand{\eureka}{\textsc{Eureka}}
\newcommand{\beam}{\textsc{Lean Beam}}
\newcommand{\model}{DeepSeek-V4.1-Flash}

\usepackage{fontawesome5}
\usepackage{xfp}
\definecolor{hfyellow}{HTML}{FFD21E}
\definecolor{hforange}{HTML}{FF9D0B}
\definecolor{hfdark}{HTML}{3A3B45}
\definecolor{linkink}{HTML}{1F4E8C}
\newcommand{\hficon}{\raisebox{-0.18em}{\begin{tikzpicture}[x=1em,y=1em]
  \fill[hfyellow] (0,0) circle (0.46);
  \fill[hfdark] (-0.16,0.08) circle (0.055); \fill[hfdark] (0.16,0.08) circle (0.055);
  \draw[hfdark, line width=0.07em, line cap=round] (-0.2,-0.1) .. controls (-0.08,-0.26) and (0.08,-0.26) .. (0.2,-0.1);
  \fill[hforange] (-0.34,-0.3) ellipse (0.15 and 0.11); \fill[hforange] (0.34,-0.3) ellipse (0.15 and 0.11);
\end{tikzpicture}}}
\newcommand{\titlelink}[3]{\href{#2}{\textcolor{linkink}{#1\,#3}}}

\title{\eureka{}: Lean Proofs \\ as a Public Research Service}

\author{%
  Debargha Ganguly$^{1}$, Vikash Singh$^{1}$, Wei Zhang$^{2}$, Joe Stubbs$^{2}$, Sreehari Sankar$^{1}$,\\
  \textbf{Mohsen Hariri$^{1}$, Weicong Chen$^{1}$, Yanyan Zhang$^{1}$, Shouren Wang$^{1}$, Wang Yang$^{1}$,}\\
  \textbf{Chaoda Song$^{1}$, Alexander Nemecek$^{1}$, Osama Zafar$^{1}$, Idan Davidovich$^{1}$, Yuting Shao$^{1}$,}\\
  \textbf{Kai Ye$^{1}$, Biyao Zhang$^{1}$, Xinpeng Li$^{1}$, Vipin Chaudhary$^{1}$}\\[4pt]
  $^{1}$Case Western Reserve University\\
  $^{2}$Texas Advanced Computing Center, The University of Texas at Austin\\[6pt]
  {\small\normalfont
  \titlelink{\faGlobe}{https://by-eureka.org}{Website}\qquad
  \titlelink{\faGithub}{https://github.com/by-eureka/eureka}{Code}\qquad
  \titlelink{\hficon}{https://huggingface.co/datasets/By-Eureka/eureka-rolloutsamples}{Recorded runs}\qquad
  \titlelink{\faEnvelope[regular]}{mailto:debargha@case.edu}{\texttt{debargha@case.edu}}}
}

\begin{document}

\maketitle

\begin{abstract}
\eureka{} is a public service that completes proofs in Lean. It runs an
open-weight model, \model{}, with no task-specific training, on research
computing at the Texas Advanced Computing Center (TACC). A proof is returned
after an independent checker confirms that the user's statements are
unchanged and replays the proof in Lean's kernel. Across eight benchmarks in competition
mathematics, graduate stochastic processes and program verification,
\eureka{} proves $1{,}896$ of $2{,}012$ entries ($94.2\%$; the $12$ Putnam~2025
problems also appear in PutnamBench). On PutnamBench it proves $670$ of $672$
problems for $\$0.17$ per problem, counting failed attempts and pricing tokens
at DeepSeek's off-peak list rates. This ties the lowest cost reported at
this level of coverage; the other systems that reach it report $\$3.52$ to
$\$74$ per problem. \eureka{} also proves all $12$ Putnam~2025 problems, all
$77$ AlgoVeri tasks (writing each program as well as its proof), $188$ of $189$
VERINA tasks and $585$ of $600$ NTP4VC verification conditions, and refutes
the remaining VERINA task and nine of the NTP4VC conditions. In the
recorded runs, the model does most of its reasoning in the first third of a
search and then tests candidate steps in Lean; model generation takes $86\%$ of single-agent run
time, and cached context accounts for $97\%$ of tokens and $20\%$ of cost. The
harness and Lean tooling are open source, and all recorded runs are
released.
\end{abstract}

\section{Introduction}
\label{sec:intro}

A proof in Lean~\citep{demoura2021lean4} is checked step by step by Lean's
kernel against the stated theorem. Lean can state both mathematical theorems
and properties of programs. Language models now write such proofs by proposing Lean
code and revising it on the checker's feedback. In the AlphaProof Nexus study,
an agent resolved nine open Erd\H{o}s problems, and a basic loop alternating
model generation with Lean verification matched those successes, at greater
cost on the hardest problems~\citep{tsoukalas2026nexus}. Several of the strongest provers use closed
models~\citep{seedprover152025,zhang2026humanizeolympiadagentsmultisession}
and are not offered as services.

\eureka{} is a public proving service.\footnote{\url{https://by-eureka.org}, in research alpha. Access requires an approved TACC account, as it runs on NSF-funded systems; email us to get set up. The code is open source for self-hosting, and contributions are welcome.} A user submits Lean statements with \li{sorry}
where a proof (or, for programs, an implementation) is wanted, and receives
a completed file. An open-weight model served from the Vista
supercomputer~\citep{taccvista} drives the search, and a shared
pool of Lean workers checks its attempts. Before a result is returned,
Comparator~\citep{comparator} rebuilds the original statements in a fresh
project, confirms they are unchanged, rejects \li{sorry} and non-standard
axioms, and replays the proof in the kernel.

\begin{figure}[t]
\centering
\includegraphics[width=\linewidth]{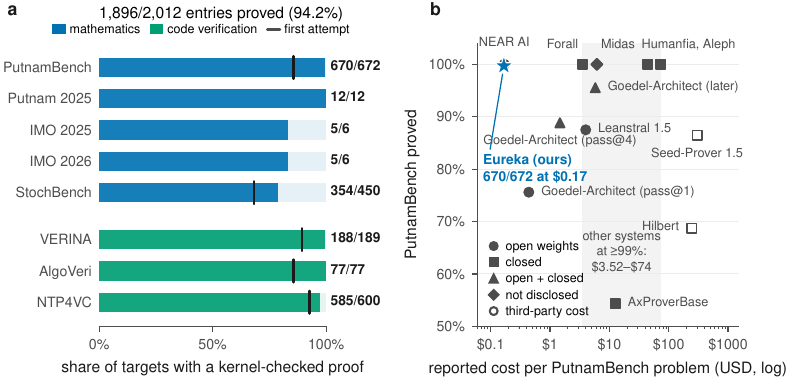}
\caption{\textbf{Coverage on eight benchmarks and PutnamBench cost.}
(a)~Share of each benchmark with a kernel-checked proof after continued runs
(bars) and on the first attempt (black ticks; omitted for the olympiad sets,
whose later runs build on earlier work). (b)~PutnamBench coverage against
reported cost per problem. \eureka{} (star) matches the lowest reported cost
(NEAR AI, $\$0.17$ at DeepSeek's earlier prices, beneath the star); the other systems at ${\geq}99\%$ report
$\$3.52$--$\$74$ (shaded). Costs follow each source's own accounting; ours is
the token charge of every run, including failures, at DeepSeek's off-peak
list rates (\cref{sec:setup}). Putnam~2025 is also part of PutnamBench.}
\label{fig:headline}
\end{figure}

We evaluated the deployed service on eight Lean benchmarks
(\cref{fig:headline}).
\begin{itemize}[leftmargin=*, itemsep=1pt, topsep=2pt]
\item With \model{} and no task-specific training, \eureka{} proves $670/672$ PutnamBench problems and $850$ of $866$
program-verification tasks (\cref{sec:results}).
\item Counting every failed attempt, PutnamBench costs $\$0.17$
per problem in token charges. This ties the lowest reported cost at this
coverage and is $21$ to $440$ times below the other systems at ${\geq}99\%$.
\item The first recorded run proves $83.4\%$ of entries;
restarting from saved work raises this to $94.2\%$, and on the olympiad sets
from $5$ to $22$ of $24$ problems.
\item Model generation takes $86\%$ of run time and Lean checking $9\%$. Over
a long run the context grows about tenfold and the cost per turn about
threefold, since cached input is priced low (\cref{sec:trajectories}).
\end{itemize}
The harness, Lean tooling and verifier are open
source~\citep{eurekacode,leanbeam,comparator}, and all recorded runs are
released on Hugging Face, with access on request~\citep{eurekarollouts}.

\section{The \eureka{} service}
\label{sec:system}

An agent harness drives the model, a shared pool of Lean workers evaluates
proof steps, and an independent verifier checks each submitted proof.

\paragraph{Workflows.}
Besides autonomous proving (\li{autoprove}, evaluated here), the platform
supports guided proving, disproof, drafting and autoformalization from
informal prompts, source review, and refactoring or golfing a working proof.

\begin{figure}[t]
\centering
\includegraphics[width=0.78\linewidth]{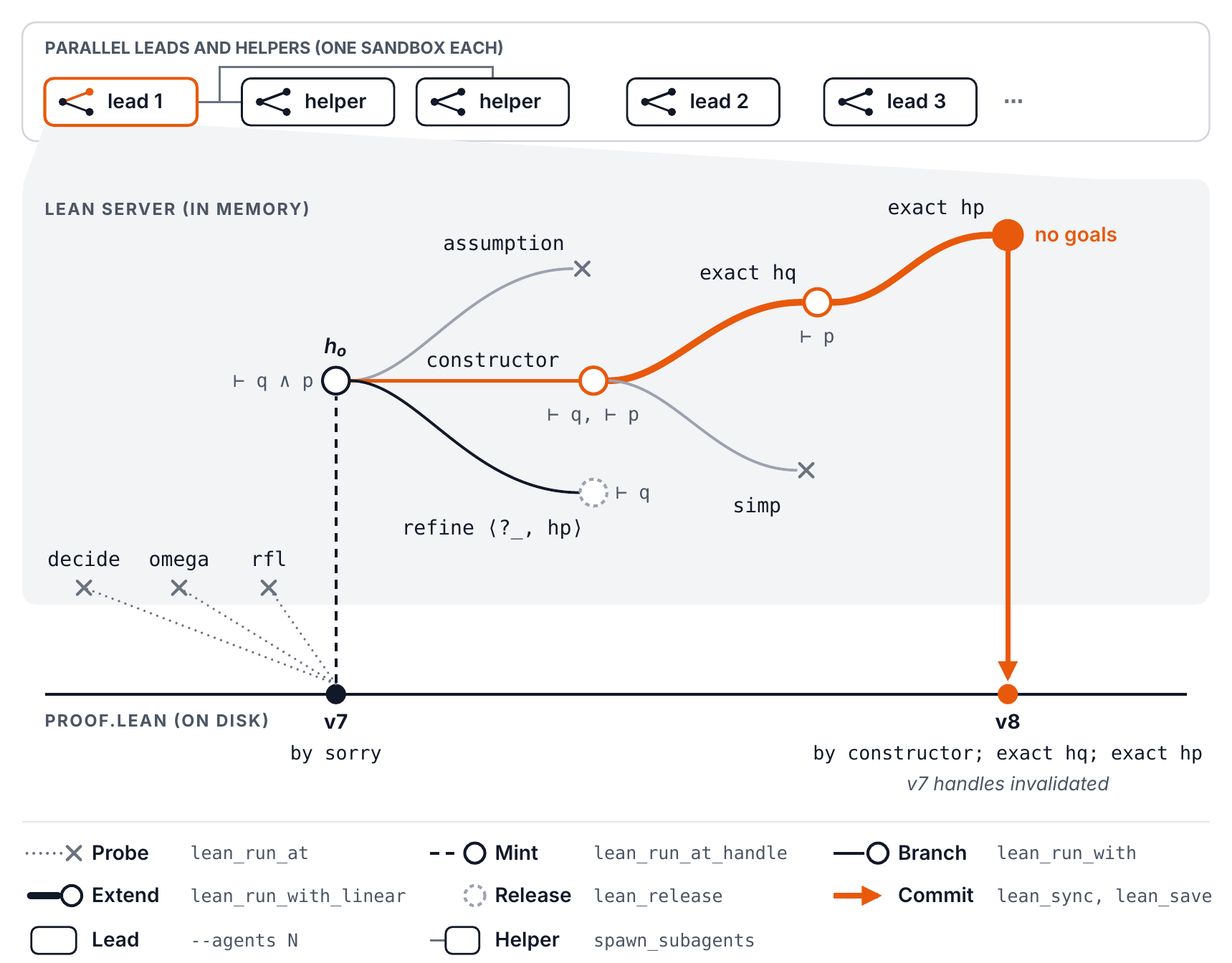}
\caption{\textbf{Proof search with \beam{}.} Top: independent leads, each with optional helpers, run in separate
sandboxes; the first verified proof is published and the other leads are
cancelled. Below: from a \li{sorry} for $q \wedge p$, one-shot probes
(dotted) test tactics without keeping state, while saved handles allow
branching (which keeps the parent) and linear extension (which consumes it).
Failed tactics ($\times$) leave no handle. The agent writes the completed
orange branch to disk, producing version v8 and invalidating v7's handles. In
the evaluation, one-shot probes and batched checks did most of the work.}
\label{fig:lean-beam}
\end{figure}

\paragraph{Agent loop.}
On each turn the harness~\citep{eurekacode} sends the agent's instructions and
conversation to the model, runs the requested tool calls and appends their
results. The agent can read project files, edit the target file and query
Lean (\cref{app:tools}). Between context compactions the conversation is only
appended to, so successive requests share a cached prefix. Agents test
candidate steps against an in-memory Lean state before editing the file
(\cref{fig:lean-beam}).

\paragraph{Helpers.}
A lead can start up to $8$ helpers, each with its own instructions and
conversation, starting from a snapshot of the lead's file. The lead
continues while helpers run and may merge parts of the files they return; only
the lead can submit.

\begin{figure}[t]
\centering
\includegraphics[width=0.88\linewidth]{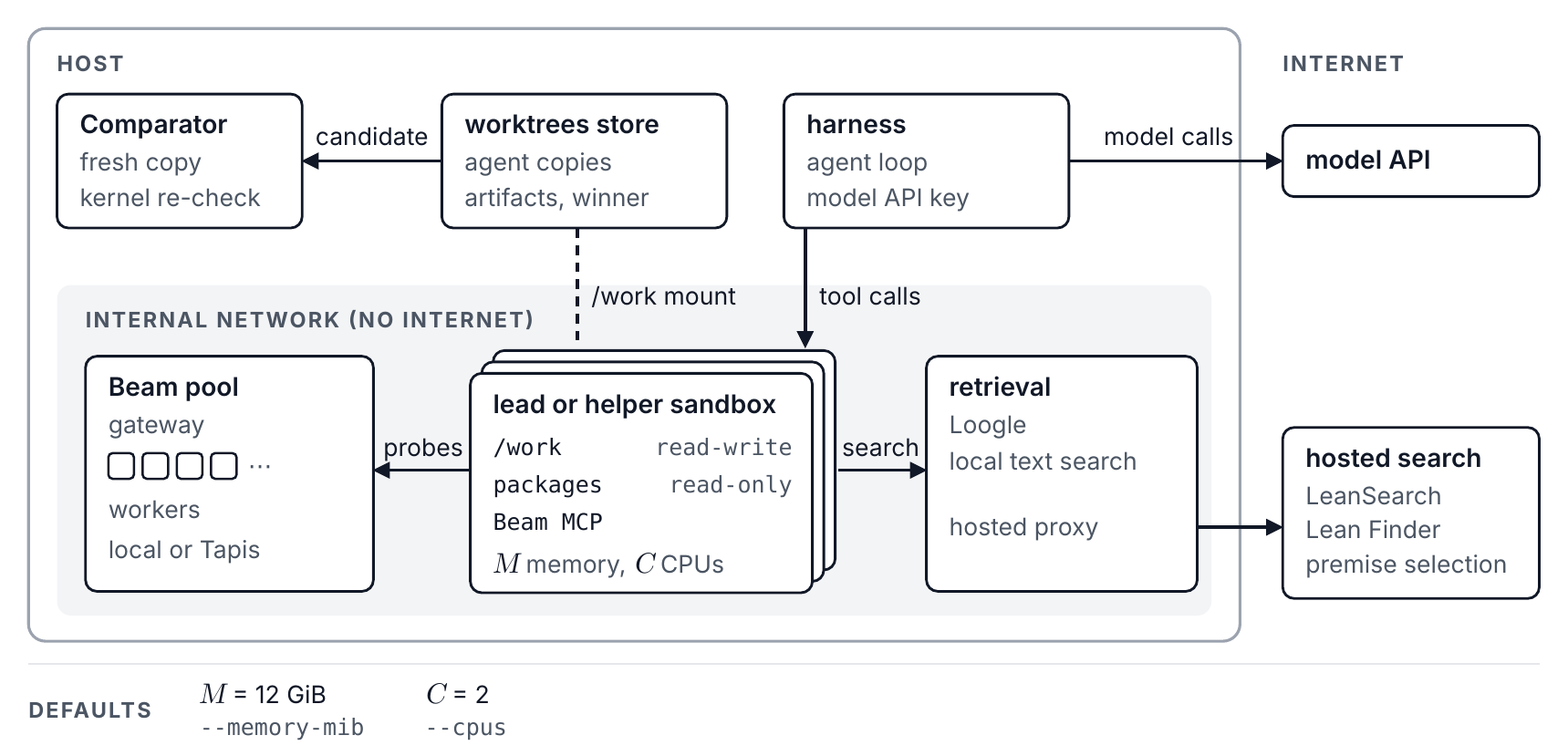}
\caption{\textbf{Isolation of agents from the host and from each other.}
The harness, model API key, result store and Comparator stay on the host.
Each lead or helper runs in its own container (memory $M$, $C$ CPUs) on an
internal network with no internet access, seeing only its own project copy
and a read-only Mathlib. Speculative Lean requests go to the shared \beam{}
pool, run locally or on Tapis; the retrieval service is the only container
with outbound access. Comparator receives only the submitted file and checks
it in a fresh copy of the trusted template.}
\label{fig:isolation}
\end{figure}

\paragraph{Verification.}
On \li{submit}, the lead's conversation ends and its helpers stop.
Comparator~\citep{comparator} builds the original and the candidate in
separate sandboxes and checks that the theorem statements and every
declaration they use are unchanged. It rejects remaining holes and any axiom
beyond \li{propext}, \li{Quot.sound} and \li{Classical.choice}, then replays
the candidate in the Lean kernel on the host (\cref{fig:isolation}).

\paragraph{Restarts and compaction.}
If a run ends without an accepted proof, a new conversation starts from the
saved file, the lead's notes and any verifier report. A conversation that
outgrows the context window is replaced by the task and a handoff summary;
this happened in $11$ runs.

\paragraph{Shared Lean elaboration.}
\beam{}~\citep{leanbeam} extends the Lean language server with speculative
elaboration at any position in a file. Its workers keep prepared projects and
Mathlib loaded, so probes do not re-import Mathlib. Each agent keeps its own
Lean session for its file, and the pool runs speculative requests for all
agents (\cref{app:pool}, \cref{fig:pool-routing}).

\section{Evaluation setup}
\label{sec:setup}

\paragraph{Benchmarks.}
We evaluate on eight Lean benchmarks with $2{,}012$ entries ($2{,}000$ distinct
problems). \emph{Competition
mathematics}: PutnamBench ($672$ problems)~\citep{putnambench2024}, Putnam~2025
($12$, also part of PutnamBench but reported separately, as other systems do),
IMO~2025 ($6$) and IMO~2026 ($6$). \emph{Graduate stochastic processes}:
StochBench ($450$ targets)~\citep{stochbench2026}, a recent benchmark on Markov
chains, martingales, renewal theory, queues, Brownian motion and stochastic
calculus. About three quarters of its targets assume, as hypotheses,
properties of objects that Mathlib does not define. \emph{Program verification}: VERINA ($189$ tasks,
proof generation)~\citep{verina2025} and NTP4VC ($600$ verification
conditions)~\citep{ntp4vc2026}, which supply the program and its
specification, and AlgoVeri ($77$)~\citep{zhao2026algoveri}, which supplies
only the specification, so the agent writes the program and proves it
correct. We ported NTP4VC and AlgoVeri to Lean~4.33.0 with the minimal edits
needed to compile. One VERINA statement omits a precondition its theorem needs; we
count it as unsolved.

\paragraph{Model and protocol.}
The evaluation uses \model{}~\citep{deepseekv41flash}, an open-weight model;
we used two checkpoints, DeepSeek-V4.1-Flash and its 1017 update, neither fine-tuned. A turn may produce up to $300{,}000$
output tokens including reasoning, agents may take up to $10{,}000$ turns,
and runs have no wall-clock limit; the agent is told the statement is provable
and to keep going until a proof is verified. Each run used a single lead. Entries without an accepted proof were
restarted from their saved work (\cref{sec:system}). The olympiad runs also
used \model{} but were made before model identifiers were logged; one
IMO~2026 history includes a run with Kimi-K3.1-Pro.

\paragraph{Scoring.}
An entry is solved when Comparator accepts a proof of its supplied statement;
refutations of a specification are reported separately. A \emph{run} is one lead
conversation with its helpers, and an \emph{attempt} is a run that called the
model. Of the $3{,}121$ runs, $598$ made no model call (infrastructure stops and
verifier-only rechecks) and are excluded from attempts and time. \emph{First attempt} means
acceptance after one model-calling run.

\paragraph{Cost and time.}
The model ran on research computing, so no costs were incurred. To compare
with systems that buy tokens, we price them at DeepSeek's off-peak list
rates for the Flash model~\citep{deepseekpricing2026}: $\$0.15$ per million uncached input
tokens, $\$0.003$ per million cached input tokens and $\$0.60$ per million
output tokens, including reasoning. Input is counted on every request,
including cached prefixes. The \emph{all-in} cost of an entry covers every run
and helper up to its first accepted proof (all runs if it was never proved);
the \emph{proof run} is the run that produced the accepted file. Proving time
sums the elapsed time of model-calling runs over the same span; helpers run
alongside their lead and are not added again. Time medians are over solved
entries; intervals are bootstrap or Wilson $95\%$ intervals over entries.

\section{Results}
\label{sec:results}

\subsection{Coverage}
\eureka{} proves $1{,}896$ of $2{,}012$ benchmark entries ($94.2\%$;
\cref{tab:ours}), including $99.7\%$ of PutnamBench, every Putnam~2025 and
AlgoVeri problem, $99.5\%$ of VERINA and $97.5\%$ of NTP4VC. The median proof
run costs $\$0.056$, and the median solved entry takes $40$ minutes all-in.
StochBench accounts for $96$ of the $116$ unsolved entries.

\begin{table}[t]
\centering
\caption{\textbf{\eureka{} with \model{} on each benchmark.} \emph{First
attempt}: proved by the first model-calling run (omitted for the olympiad
sets, whose later runs built on earlier work). \emph{Proof run}: the run that
produced the accepted file. \emph{All-in}: every run up to the first accepted
proof, including failures. Proof-run cost and both times are medians over
solved entries; all-in cost is the mean over all entries, solved or not, at
the rates in \cref{sec:setup}. The \emph{All} row's first-attempt rate
includes the olympiad sets. Green shading scales with coverage;
\textcolor{goodgreen}{$\blacktriangle$} gives the gain in points from restarts.}
\label{tab:ours}
\footnotesize
\setlength{\tabcolsep}{3.4pt}
\begin{tabular}{lrrrrrrr}
\toprule
& & & & \multicolumn{2}{c}{Cost (USD)} & \multicolumn{2}{c}{Median time (min)} \\
\cmidrule(lr){5-6}\cmidrule(lr){7-8}
Benchmark & $n$ & Proved & First attempt & Proof run & All-in / entry & Proof run & All-in \\
\midrule
PutnamBench  & 672 & \heat{40}{\textbf{670 (99.7\%)}} & 85.7\%\gain{14.0} & 0.070 & 0.168 & 53 & 59 \\
Putnam 2025  & 12 & \heat{40}{\textbf{12 (100\%)}} & \na & 0.162 & 0.316 & 52 & 105 \\
IMO 2025     & 6 & \heat{23}{5 (83.3\%)} & \na & 0.325 & 1.123 & 86 & 144 \\
IMO 2026     & 6 & \heat{23}{5 (83.3\%)} & \na & 0.233 & 2.559 & 29 & 105 \\
StochBench   & 450 & \heat{19}{354 (78.7\%)} & 68.2\%\gain{10.5} & 0.069 & 0.205 & 48 & 51 \\
VERINA       & 189 & \heat{40}{\textbf{188 (99.5\%)}} & 89.4\%\gain{10.1} & 0.033 & 0.056 & 18 & 19 \\
AlgoVeri     & 77 & \heat{40}{\textbf{77 (100\%)}} & 85.7\%\gain{14.3} & 0.051 & 0.091 & 34 & 49 \\
NTP4VC       & 600 & \heat{38}{\textbf{585 (97.5\%)}} & 92.7\%\gain{4.8} & 0.045 & 0.098 & 26 & 27 \\
\midrule
\ours All    & 2{,}012 & \heat{34}{\textbf{1{,}896 (94.2\%)}} & 83.4\%\gain{10.8} & 0.056 & 0.153 & 37 & 40 \\
\bottomrule
\end{tabular}
\end{table}

\paragraph{Competition mathematics.}
The unsolved problems are PutnamBench 1979~B5 and 2021~A6, IMO~2025~Q6 and
IMO~2026~Q3 (\cref{tab:math}). Among published systems, HOA reports more
PutnamBench proofs, and EasyProver and Numina-Lean-Agent also prove all of
Putnam~2025; these three use closed models. The other open-weight systems reach
at most $87.5\%$ in their papers. On the leaderboard, the open-weight entries
reach $97.9\%$ and $100\%$ with different inputs, namely supplied informal
proofs and repaired statements (\cref{tab:math};
\citealp{putnambenchleaderboard2026,nearai2026}).

\begin{table}[t]
\centering
\caption{\textbf{Reported results in formal mathematics.} PutnamBench counts are out of $672$; leaderboard entries checked 13
November 2026. Grey text after a system names its model where the system name
does not; small grey numbers give the reported attempt budget. Costs keep each
source's accounting; $^\ast$third-party estimate (Goedel-Architect for
Hilbert, Leanstral~1.5 for Seed-Prover~1.5). $^\dagger$No-answer track;
$^\ddagger$informal reference proofs supplied; $^\S$two repaired
formalizations; cost at DeepSeek's earlier prices ($\$0.49$ at current off-peak rates). \emph{Mixed}: open and closed models; \emph{n/d}: not
disclosed. Green shading scales with PutnamBench coverage; \textcolor{badred}{red} figures
give each reported cost as a multiple of ours ($\$0.168$).}
\label{tab:math}
\footnotesize
\setlength{\tabcolsep}{1.3pt}
\setlength{\aboverulesep}{0pt}\setlength{\belowrulesep}{0pt}\setlength{\extrarowheight}{0.4ex}
\begin{tabular}{llccccl}
\toprule
System & Weights & PutnamBench & Putnam'25 & IMO'25 & IMO'26 & Cost \\
\midrule
\multicolumn{7}{l}{\emph{Published systems}} \\
Goedel-Prover-V2-32B~\citep{goedelproverv2} & \wopen & \heat{0}{12.8\budget{@184}} & \na & \na & \na & \na \\
LongCat-Flash-Prover~\citep{longcat2026} & \wopen & \heat{1}{41.5\budget{@118}} & \na & \na & \na & \na \\
Seed-Prover~\citep{seedprover2025} & \wclosed & \heat{6}{49.0} & \na & \best{5/6} & \na & \na \\
AxProverBase~\citep{requena2026minimalagent}\budget{Claude Opus 4.5} & \wclosed & \heat{10}{54.3} & \na & \na & \na & \$12.60\xours{75} \\
Hilbert~\citep{hilbert2025}\budget{Gemini 2.5 Pro} & \wclosed & \heat{19}{68.8\budget{@1840}} & \na & \na & \na & \$244$^{\ast}$\xours{1.5k} \\
Goedel-Architect~\citep{chung2026goedel}\budget{DeepSeek-V4-Flash} & \wopen & \heat{24}{75.6\budget{@1}} & \na & \na & \na & \$0.44\xours{2.6} \\
\quad + NL guidance\budget{+ Gemini 3.1 Pro} & \wmixed & \heat{33}{88.8\budget{@4}} & 11/12 & 4/6 & \na & \$1.47\xours{8.8} \\
Seed-Prover 1.5~\citep{seedprover152025} & \wclosed & \heat{31}{86.5} & 11/12 & \best{5/6} & \na & \$300$^{\ast}$\xours{1.8k} \\
Leanstral 1.5~\citep{mistralLeanstral15} & \wopen & \heat{32}{87.5\budget{@8}} & \na & \na & \na & \$4\xours{24} \\
Numina-Lean-Agent~\citep{numinaleanagent2026}\budget{Claude Opus 4.5} & \wclosed & \na & \best{12/12} & \na & \na & \na \\
HOA~\citep{zhang2026humanizeolympiadagentsmultisession}\budget{GPT-5.6 Sol} & \wclosed & \heat{40}{\best{100.0}} & \na & \na & \best{6/6} & \na \\
AxiomProver~\citep{axiomimo2026} & \wnd & \na & \na & \na & \best{6/6} & \na \\
EasyProver~\citep{ganguly2026easyprover}\budget{Fable 5} & \wclosed & \heat{40}{99.6} & \best{12/12} & \na & \best{6/6} & \na \\
\midrule
\multicolumn{7}{l}{\emph{PutnamBench leaderboard entries}~\citep{putnambenchleaderboard2026}} \\
Goedel-Architect, later run & \wmixed & \heat{37}{95.5} & \na & \na & \na & \$5.80\xours{35} \\
Mistral Prover$^{\ddagger}$ & \wopen & \heat{39}{97.9} & \na & \na & \na & \na \\
Forall (Astrio) & \wclosed & \heat{40}{\best{100.0}} & \na & \na & \na & \$3.52\xours{21} \\
Midas Prover$^{\dagger}$ & \wnd & \heat{40}{\best{100.0}} & \na & \na & \na & \$6.17\xours{37} \\
Humanfia & \wclosed & \heat{40}{\best{100.0}} & \na & \na & \na & \$44.50\xours{265} \\
Aleph Prover & \wclosed & \heat{40}{\best{100.0}} & \na & \na & \na & \$74\xours{440} \\
NEAR AI$^{\S}$\budget{DeepSeek-V4} & \wopen & \heat{40}{\best{100.0}} & \na & \na & \na & \best{\$0.17} \\
\midrule
\ours \eureka{} (ours)\budget{\model{}} & \wopen & \heat{40}{99.7} & \best{12/12} & \best{5/6} & 5/6 & \best{\$0.17} \\
\bottomrule
\end{tabular}
\end{table}

\paragraph{Program verification.}
\eureka{} proves $188/189$ VERINA tasks, $77/77$ AlgoVeri tasks, writing each
implementation as well as its proof, and $585/600$ NTP4VC conditions
(\cref{tab:code} in \cref{app:more}). The general models and neural provers
evaluated by \citet{goedelcodeprover2026} and \citet{ntp4vc2026} reach at most
$24.3\%$ on VERINA, $23.4\%$ on AlgoVeri and $3.0\%$ on NTP4VC; the
task-trained Goedel-Code-Prover reaches $68.8\%$ and $62.3\%$ with
implementations supplied. EasyProver proves $588$ NTP4VC conditions and
refutes nine. \eureka{} refutes the remaining VERINA task, whose statement
lacks a needed precondition, and nine NTP4VC conditions. In total it resolves all
$189$ VERINA tasks and $594$ of $600$ NTP4VC conditions ($99.0\%$). The ten
refutation runs cost $\$0.26$ together (median $14$ minutes); counting them and
all earlier attempts, a resolved NTP4VC entry costs $\$0.085$ on average.

\paragraph{Graduate stochastic processes.}
On StochBench, \eureka{} proves $354$ of $450$ targets ($78.7\%$); the
benchmark's published agent proves $157$ under a $15$-minute
limit~\citep{stochbench2026}. Of the $114$ \emph{direct} targets, stated with
Mathlib or shared definitions, it proves $111$ ($97.4\%$). Of the $336$
\emph{abstracted} targets, which assume properties of objects Mathlib does not
yet define, it proves $243$ ($72.3\%$). Of the $96$ unsolved targets, $93$
are abstracted, which suggests that missing library definitions account for
most of the gap. A one-hour refutation search on each unsolved target refutes
two abstracted targets. No refutation was found for the rest, consistent with
their being correctly stated and hard to prove.

\subsection{Cost, budgets and restarts}
\paragraph{Total cost.}
Through each entry's first accepted proof, the evaluation costs $\$307.24$,
or $\$0.153$ per entry: $58\%$ for proof-producing runs, $15\%$ for earlier
attempts on entries later proved, and $27\%$ for the $116$ unsolved entries.
PutnamBench accounts for $\$112.60$, or $\$0.168$ per problem
(\cref{fig:headline}b). NEAR AI, which also uses DeepSeek models, reports the
same $\$0.17$ mean at DeepSeek's earlier prices, or $\$0.49$ at the off-peak rates
used here; the other systems at ${\geq}99\%$ report $\$3.52$ to
$\$74$. With cached input priced ten times higher, PutnamBench would cost $\$0.45$
per problem, below every other system at full coverage.
Because sources compute cost differently, these comparisons are between
reported prices.

\begin{figure}[t]
\centering
\includegraphics[width=0.84\linewidth]{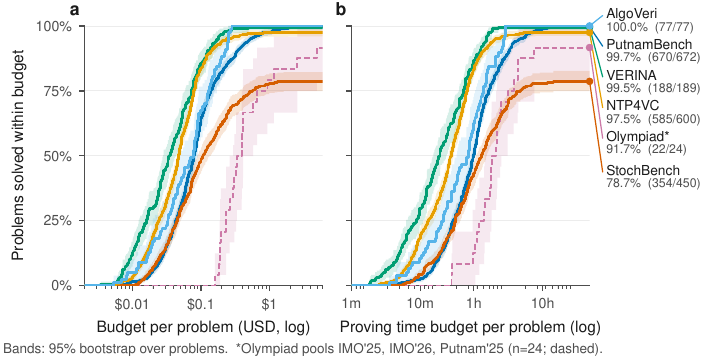}
\caption{\textbf{Coverage within a cost or time budget.} Share of each benchmark proved within a
per-entry budget of (a)~all-in token cost or (b)~proving time, counting every
run; unsolved entries stay in the denominator. Bands are $95\%$ bootstrap
intervals. Olympiad pools IMO~2025, IMO~2026 and Putnam~2025 ($24$ entries).}
\label{fig:scaling}
\end{figure}

\paragraph{Coverage within a budget.}
\Cref{fig:scaling} counts every run and keeps unsolved entries in the
denominator. Ten cents per entry proves $86\%$ of VERINA, $84\%$ of NTP4VC,
$68\%$ of AlgoVeri and $64\%$ of PutnamBench. Reaching $80\%$ costs
$\$0.08$--$\$0.09$ on VERINA and NTP4VC, $\$0.15$ on AlgoVeri, $\$0.18$ on
PutnamBench and about $\$1$ on the olympiad sets. Within $30$ minutes,
\eureka{} proves $64\%$ of VERINA, $54\%$ of NTP4VC and $25\%$ of PutnamBench.
StochBench levels off at $78.7\%$. \Cref{fig:scaling-success} in
\cref{app:more} repeats the curves for the proof run alone.

\paragraph{Restarts.}
The first attempt proves $1{,}679$ of $2{,}012$ entries ($83.4\%$); restarting
from saved work raises this to $94.2\%$ (\cref{fig:headline}a). On the
olympiad sets, $5$ of $24$ problems were proved by their first recorded run
and $22$ in the end. The mean all-in cost per solved entry is
$1.1$ times the mean proof run's on VERINA, $1.3$ on AlgoVeri, $1.4$ on
PutnamBench and $1.7$ on NTP4VC, and $5.3$ on the olympiad sets, about $\$1.18$ per solved problem (\cref{fig:retries}). Since the model reasons most at the start of
a run (\cref{sec:trajectories}), a restart may resume planning from the
partial proof with a shorter context; we did not test this directly.

\paragraph{Distributions.}
On code verification the 99th-percentile cost is $3.4$ to $8.2$ times the median
(\cref{fig:atlas} in \cref{app:more}). PutnamBench has the longest tail: the median solve costs
$\$0.08$ and takes $59$ minutes, the 99th percentile $18$ times as much and
$9.3$ hours.

\section{Analysis of the recorded runs}
\label{sec:trajectories}

Closed-model APIs return at most a summary of the model's reasoning. Because
\eureka{} serves an open-weight model itself, its records include the full
reasoning text of every model call, along with tool calls and proof files. We analyze the
$1{,}896$ solved entries up to their first accepted proof: $2{,}738$ runs ($2{,}224$ of
them calling the model) with $103{,}674$ model calls and $114{,}928$ tool calls,
all of which are released.

\subsection{Reasoning and tokens}
\begin{figure}[t]
\centering
\includegraphics[width=0.9\linewidth]{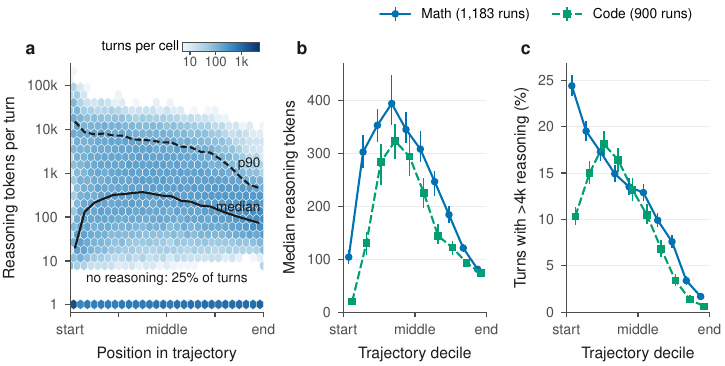}
\caption{\textbf{Reasoning is concentrated early in a run.} Lead conversations with at least ten turns ($86{,}229$ turns in
$2{,}083$ runs), position scaled to $[0,1]$. (a)~Density of reasoning tokens per
turn, with median and 90th percentile; the bottom row holds the $25\%$ of turns
with no reasoning. (b)~Median reasoning per tenth of a run (bootstrap $95\%$
intervals). (c)~Share of turns above $4{,}000$ reasoning tokens (Wilson $95\%$
intervals).}
\label{fig:thinking}
\end{figure}

\paragraph{Reasoning by position in a run.}
\Cref{fig:thinking} aligns $2{,}083$ lead conversations of at least ten turns by
relative position. On mathematics, median reasoning per turn rises to about
$390$ tokens a third of the way through, then falls fivefold to about $80$ in
the last tenth. Turns with more than $4{,}000$ reasoning tokens are $24\%$ of
the first tenth and under $2\%$ of the last. On code verification, such turns
peak at $18\%$ in the third tenth and
fall below $1\%$. With each run's last three turns removed, the mathematics peak is still $4.3$
times the final tenth. A quarter of all turns contain tool calls and no
reasoning.

\begin{figure}[t]
\centering
\includegraphics[width=0.84\linewidth]{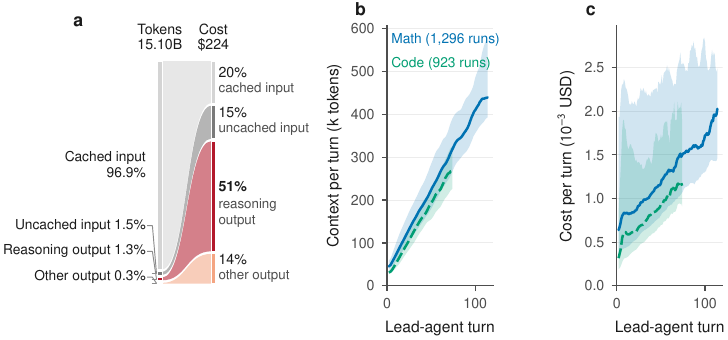}
\caption{\textbf{Token counts and cost by token type.}
Searches on solved entries. (a)~Share of tokens (left) and of cost (right) by
type; cached prefixes are counted on every request. (b)~Lead context length
and (c)~cost per turn (median and interquartile range, five-turn smoothing;
curves end where fewer than $50$ runs remain).}
\label{fig:tokens}
\end{figure}

\paragraph{Tokens and cost.}
Solved entries used $15.1$ billion tokens, priced at $\$223.78$
(\cref{fig:tokens}). Cached context is $96.9\%$ of tokens and $19.6\%$ of
cost; reasoning is $1.3\%$ of tokens and $51.3\%$ of cost. Over a long
mathematics run the lead's context grows from about $45$k tokens to $440$k by
turn $115$, while the cost per turn only triples, because most of the context is cached
and priced low.

\subsection{Time and tools}
\begin{figure}[t]
\centering
\includegraphics[width=0.7\linewidth]{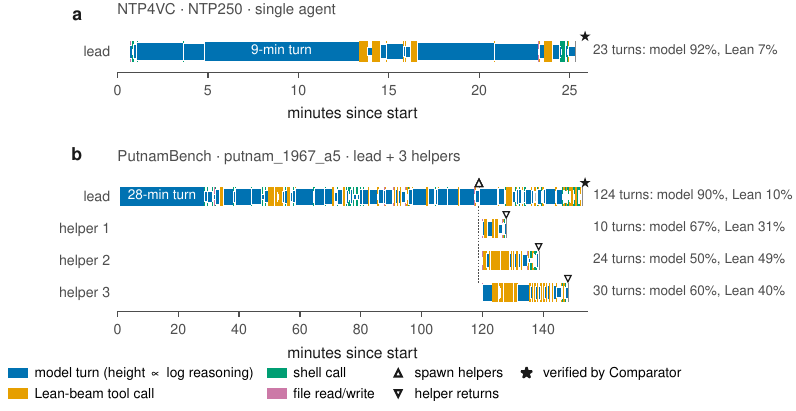}
\caption{\textbf{Timelines of two accepted searches.} (a)~NTP4VC NTP250, one
agent. (b)~PutnamBench 1967~A5, a lead and three helpers. Bar height encodes
log reasoning tokens per model call; percentages give each agent's share of
active time in model calls and Lean. The longest model calls last $9$ and
$28$ minutes and carry $17$k and $57$k reasoning tokens.}
\label{fig:anatomy}
\end{figure}

\paragraph{Time.}
In $2{,}044$ single-agent runs on solved entries, model requests take
$85.8\%$ of elapsed time and \beam{} calls $8.8\%$. \Cref{fig:anatomy} shows two
searches. In an NTP4VC solve, $23$ model calls fill $92\%$ of the agent's
active time, between short Lean checks. In a PutnamBench solve, the lead
starts three helpers after $119$ minutes and keeps working; the helpers spend
$31$--$49\%$ of their time in Lean, against $10\%$ for the lead.

\begin{figure}[t]
\centering
\includegraphics[width=0.8\linewidth]{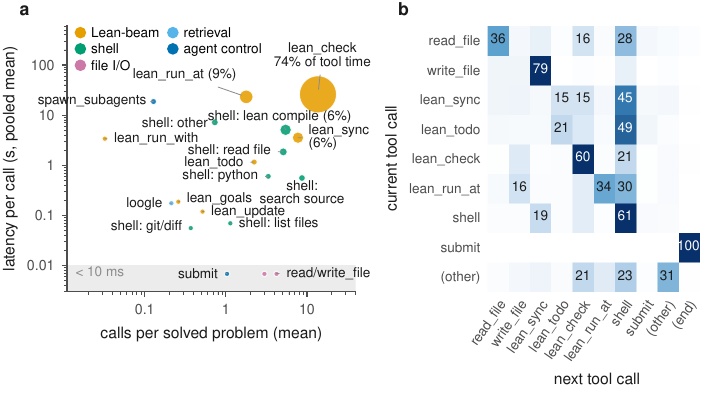}
\caption{\textbf{Tool use on solved entries.} (a)~Calls per solved entry against mean latency; bubble area is
the share of tool time, and shell calls are split by command. (b)~Probability
of the lead's next tool given the current one ($96{,}327$ transitions); cells
of $15\%$ or more are labelled.}
\label{fig:tools}
\end{figure}

\paragraph{Tool use.}
After writing the file, the lead reloads it in Lean $79\%$ of the time; a
\li{lean\_check} is followed by another $60\%$ of the time; and a shell call,
the most frequent tool, by another $61\%$ of the time, as the agent searches
Mathlib and reads files (\cref{fig:tools}). \li{lean\_check}, which elaborates
up to $32$ candidates at once, averages $26$ seconds a call and accounts for
$74\%$ of tool time; compiling Lean from the shell takes $6\%$. The agent made
$3{,}381$ one-shot \li{lean\_run\_at} probes but continued a saved proof state
only $61$ times, and used retrieval in $11\%$ of runs. Saved handles
(\cref{fig:lean-beam}) were rarely used.

\subsection{Tactics, helpers and proofs}
\begin{figure}[t]
\centering
\includegraphics[width=0.84\linewidth]{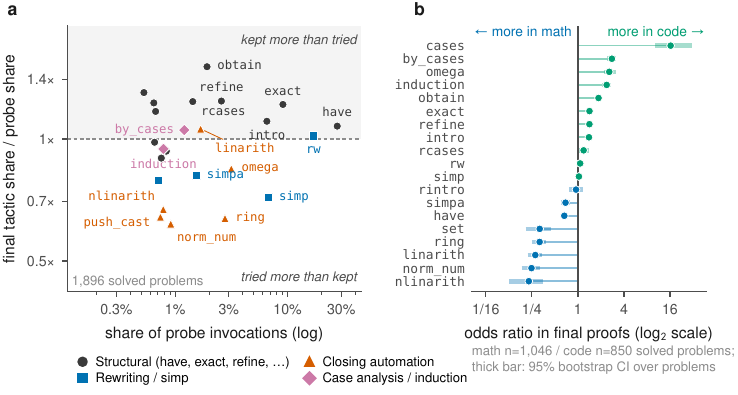}
\caption{\textbf{Tactics in probes and in accepted proofs.}
(a)~Each tactic's share of tactics in accepted proofs divided by its share in
probes (a comparison of compositions). (b)~Odds ratio of each tactic's use in
code-verification proofs ($n=850$) versus mathematics proofs ($n=1{,}046$),
with bootstrap $95\%$ intervals.}
\label{fig:tactics}
\end{figure}

\paragraph{Tactics.}
Relative to their share of probes, structural tactics are more common in
accepted proofs, \li{obtain} at $1.5$ times and \li{refine}, \li{rcases} and
\li{exact} at $1.2$, while \li{norm\_num}, \li{ring}, \li{push\_cast} and
\li{nlinarith} appear at $0.6$--$0.7$ times (\cref{fig:tactics}a). Code proofs use \li{cases} about $16$ times as often as mathematics
proofs, and \li{by\_cases}, \li{omega} and \li{induction} two to three times
as often; mathematics proofs use \li{nlinarith}, \li{norm\_num},
\li{linarith} and \li{ring} three to four and a half times as often
(\cref{fig:tactics}b).

\begin{figure}[t]
\centering
\includegraphics[width=0.74\linewidth]{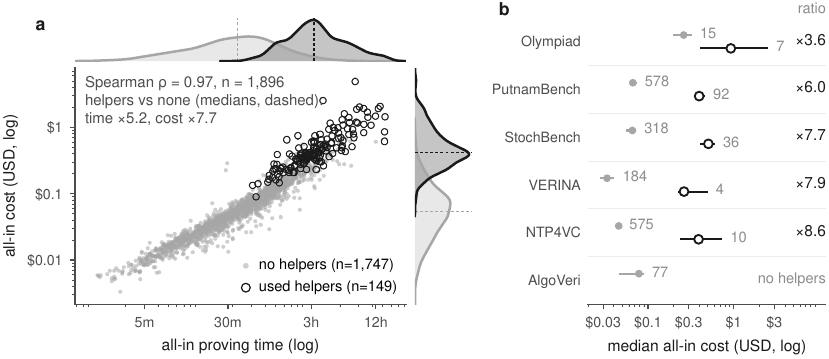}
\caption{\textbf{Cost, time and helper use.}
(a)~All-in cost against proving time for every solved entry, with marginal
densities; hollow markers are solves that used helpers ($n=149$).
(b)~Median cost per benchmark with and without helpers (bootstrap $95\%$
intervals; counts shown). AlgoVeri searches never used helpers.}
\label{fig:effort}
\end{figure}

\paragraph{Helpers.}
Of $1{,}896$ solves, $149$ ($8\%$) used helpers, mostly on PutnamBench
($92$) and StochBench ($36$), never on AlgoVeri (\cref{fig:effort}). Their
median time is $5.2$ times and median cost $7.7$ times that of other solves,
and $3.6$ to $8.6$ times within each benchmark. The lead starts helpers after a
search has already run for some time, which places them in the longer
searches. Cost and time are strongly correlated (Spearman $0.97$), so a time limit
effectively also limits cost.

\subsection{Interpretation}
\begin{itemize}[leftmargin=*, itemsep=1pt, topsep=2pt]
\item Long reasoning early in a run, a quarter of turns without reasoning,
and runs of batched checks are consistent with a search that forms a plan in
a few long turns and then tests local variants in Lean. The first third of a run holds $59\%$ of
reasoning tokens on mathematics and $50\%$ on code, and reasoning is $51\%$ of
cost, so these planning turns account for much of what a search costs.
\item Most tool time is spent checking candidate steps in Lean; retrieval
and saved proof states are rarely used.
\item With no fixed per-problem budget, spending is highest on the hardest
problems: helpers appear in $8\%$ of solves, restarts are concentrated on the
olympiad sets, and competition problems have a long cost tail. A shared
service could cap the tail per user while leaving typical searches unlimited.
\end{itemize}

\section{Related work}
\label{sec:related}

\paragraph{Provers.}
Language-model provers are typically trained on proof corpora and search with
a learned policy, generating single steps~\citep{polu2020gpt} or whole
proofs~\citep{first2023baldur}, with premise selection~\citep{leandojo2023}
and tree search~\citep{lample2022hypertree}; Goedel-Prover-V2~\citep{goedelproverv2},
DeepSeek-Prover-V2~\citep{ren2025deepseek} and
Seed-Prover~\citep{seedprover2025,seedprover152025} add expert
iteration~\citep{polu2022curriculum}, reinforcement learning and large inference
budgets. A second line places a general model in an agent loop with Lean as a
tool~\citep{yao2023react}: Hilbert~\citep{hilbert2025},
Goedel-Architect~\citep{chung2026goedel},
Numina-Lean-Agent~\citep{numinaleanagent2026},
AxProverBase~\citep{requena2026minimalagent} and
EasyProver~\citep{ganguly2026easyprover}; for code,
Goedel-Code-Prover~\citep{goedelcodeprover2026} trains a goal-decomposition
policy. \eureka{} belongs to the second line and uses no task-specific
training. It runs as a public service on an open model, and we report its cost and
release its reasoning traces together with its accuracy.

\paragraph{Lean interfaces and verification.}
LeanDojo~\citep{leandojo2023} and \li{lean-lsp-mcp}~\citep{lean-lsp-mcp}
connect an agent to a local Lean server; \beam{} serves speculative elaboration
from a shared pool. Comparator~\citep{comparator}, also used by EasyProver,
blocks weakened statements, leftover \li{sorry}s and added
axioms~\citep{baker2025monitoring,gamingverifiers2026}.

\section{Discussion and conclusion}
\label{sec:discussion}
\label{sec:limitations}

\paragraph{Running a proving service.}
Inference capacity is the main resource to provision, since model requests
take $86\%$ of run time. With a shared Lean pool that keeps Mathlib loaded,
checking takes under a tenth of the time, and append-only conversations with
prefix caching keep long searches inexpensive. Our dollar
figures apply DeepSeek's off-peak list prices to recorded token counts; at ten
times the cached-input price the evaluation would cost $\$899.90$ instead of
$\$307.24$, or $\$0.45$ per entry across all benchmarks. Hardware and energy are outside these prices.

\paragraph{Evaluating the whole system.}
The effect of one component of an agent depends on the model, the other
components and the budget~\citep{ganguly2026evaluationagent}, so results from
ablating one component under one model may not transfer. Attributing gains
requires crossed, budget-matched comparisons. We
evaluate the deployed system end to end and release every trajectory for such
comparisons.

\paragraph{Scope.}
Comparisons use other systems' reported results, whose benchmark versions,
inputs and budgets differ. We ported two benchmarks to a newer Lean, counted
refutations separately from proofs, and used two checkpoints of \model{}; the olympiad sets are small. Static
benchmarks may overlap with training data, which could raise coverage; the
2025--2026 competition sets and StochBench, released in 2026, are the least
exposed, and \eureka{} proves $97.4\%$ of StochBench's directly stated targets.

\paragraph{Conclusion.}
An open-weight model on public research computing proves nearly all of
PutnamBench and of three program-verification benchmarks at under $20$ cents per
problem, and we release its recorded runs, including the full reasoning
traces. \eureka{} is
open to users at \href{https://by-eureka.org}{by-eureka.org}.

\label{sec:conclusion}
\section*{Acknowledgments}
The service and the evaluation run on resources of the Texas Advanced Computing Center (TACC) at The University of Texas at Austin, including Vista. This research was supported in part by NSF Awards 2117439 and 2112606.

\bibliography{ref}
\bibliographystyle{plainnat}

\appendix
\section{Tools available to the agent}
\label{app:tools}

\Cref{tab:tools} lists the tools the agents used. The \beam{} server exposes a
few further navigation tools (hover, definitions, references, symbols, code
actions) that the agent rarely calls, and tool output is truncated to
$64$~KiB, keeping its head and tail. Across lead conversations on solved
entries, the agent made $39{,}914$ shell calls,
$20{,}615$ \li{lean\_check} calls, $12{,}916$ reloads with \li{lean\_sync},
$6{,}487$ file reads and $4{,}901$ file writes. Shell calls are the most
numerous, mostly short searches and file reads, while \li{lean\_check} takes
the most tool time (\cref{fig:tools}). Each \li{lean\_check} call can test up to
$32$ candidates.

\begin{table}[!htbp]
\centering
\caption{\textbf{Tools available to the agents, with measured use.} Calls per
solved entry, mean latency per call and share of summed tool time, over all
agents on the $1{,}896$ solved entries; colors match \cref{fig:tools}. Tool
time excludes waiting for helpers and overlaps across agents.}
\label{tab:tools}
\footnotesize
\setlength{\tabcolsep}{4pt}
\renewcommand{\arraystretch}{1.12}
\begin{tabularx}{\linewidth}{@{}>{\raggedright\arraybackslash}p{3.0cm}>{\raggedright\arraybackslash}Xrrl@{}}
\toprule
Tool & What it does & Calls/solve & Mean (s) & Tool time \\
\midrule
\tcat{catlean}{Lean (\beam{} pool)}
\li{lean\_check} & Elaborates up to $32$ commands or tactic blocks against the current file, concurrently, without writing to disk & 13.7 & 26.0 & \tshare{catlean}{74.2} \\
\li{lean\_run\_at} & Elaborates one block at a file position without changing the file; returns messages and proof state & 1.8 & 23.2 & \tshare{catlean}{8.6} \\
\li{lean\_sync},\newline \li{lean\_update} & Reload the file in Lean, with or without waiting for diagnostics & 8.3 & 3.3 & \tshare{catlean}{5.8} \\
\li{lean\_todo} & Lists \li{sorry}s, holes and errors in a range & 2.2 & 1.2 & \tshare{catlean}{0.5} \\
\li{lean\_goals} & Shows the goals before or after a position & 0.3 & 0.2 & \tsmall \\
\li{lean\_run\_at\_handle},\newline \li{lean\_run\_with} & As \li{lean\_run\_at}, but save a proof-state handle or continue from one & 0.06 & 5.3 & \tshare{catlean}{0.1} \\
\li{lean\_save} & Writes a compiled checkpoint of the file & 0.04 & 7.9 & \tshare{catlean}{0.1} \\
\tcat{catshell}{Shell}
\li{shell} & Runs a command in the container (default limit $300$~s, at most $1{,}800$~s); output cut to $16$~KiB at each end & 24.8 & 2.0 & \tshare{catshell}{10.3} \\
\tcat{catfile}{Files}
\li{read\_file} & Reads a project file with line numbers & 4.2 & ${<}0.01$ & \tsmall \\
\li{write\_file} & Replaces the target file; any other path is refused & 3.0 & ${<}0.01$ & \tsmall \\
\tcat{catretr}{Retrieval (optional)}
\li{mcp\_retrieval\_*} & Loogle over a local Mathlib index, LeanSearch and a local text search & 0.3 & 0.2 & \tsmall \\
\tcat{catagent}{Agent control}
\li{spawn\_subagents} & Starts up to $8$ helpers (lead only) & 0.13 & 18.7 & \tshare{catagent}{0.5} \\
\li{wait\_subagents} & Collects finished helpers (lead only) & 0.49 & 23.0 & \multicolumn{1}{c}{\na} \\
\li{submit} & Ends the lead's run and sends the file to Comparator & 1.0 & ${<}0.01$ & \tsmall \\
\li{finish} & Ends a helper's run and returns its file to the lead & 0.23 & ${<}0.01$ & \multicolumn{1}{c}{\na} \\
\bottomrule
\end{tabularx}
\end{table}

\section{The shared Lean pool}
\label{app:pool}

\begin{figure}[!htbp]
\centering
\includegraphics[width=0.9\linewidth]{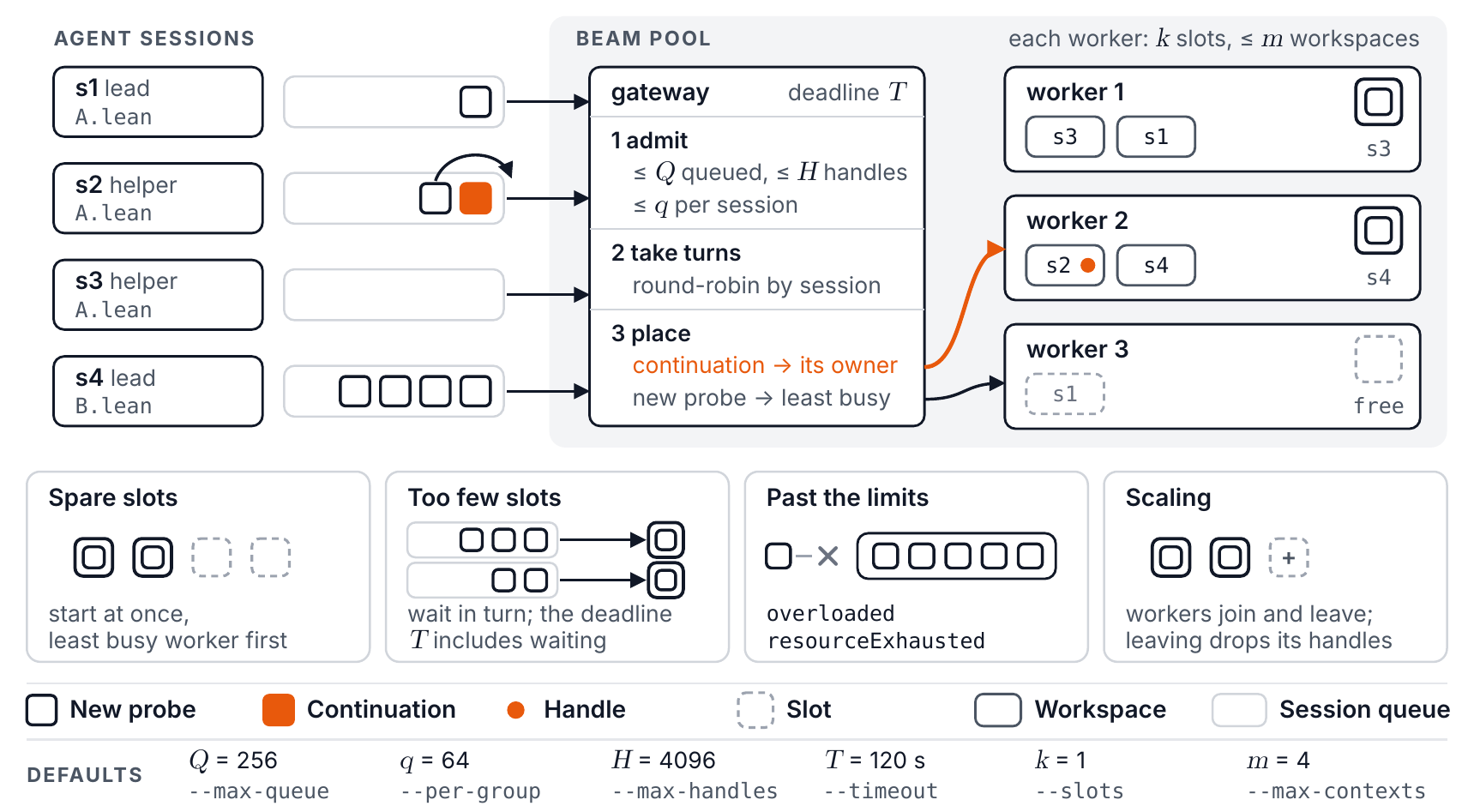}
\caption{\textbf{Request routing in the \beam{} pool.} Top: the gateway serves
session queues round-robin; a continuation returns to the worker holding its
handle (s2 waits for worker~2), while new probes go to the least busy worker
with the same prepared project. Bottom: spare slots run requests at once,
contention queues them, and requests beyond the queue or handle limits are
rejected. Added workers join after discovery.}
\label{fig:pool-routing}
\end{figure}

Each agent keeps its own Lean session for synchronizing and inspecting its
file, while the \beam{} pool executes speculative requests for all agents
(\cref{fig:pool-routing}). Project files move by content hash, so a worker
receives only the bytes it lacks. Each worker limits its session workspaces
and evicts the least recently used idle one when necessary, and the gateway
rediscovers workers every second, so capacity can be added or removed while
agents run. Queue overflow returns \li{overloaded} and exhausting the handle
limit returns \li{resourceExhausted}; the queue, handle, deadline, slot and
workspace limits are configurable. A probe reuses an environment in which Mathlib is already
loaded, and Lean checking stays under a tenth of single-agent run time
(\cref{sec:trajectories}) even though \li{lean\_check} is the agent's most
time-consuming tool.

\section{Additional results}
\label{app:more}

\subsection{Program-verification baselines}
\Cref{tab:code} gives the baselines behind the program-verification
comparison in \cref{sec:results}. The general-purpose models, several of them
frontier closed models, prove $10.6$--$19.6\%$ of VERINA and $5.2$--$18.2\%$ of
AlgoVeri at pass@128, and every model evaluated on NTP4VC proves at most
$3.0\%$ of its conditions. Neural provers trained for mathematics do little
better on code ($15.9$--$24.3\%$ on VERINA). The task-trained
Goedel-Code-Prover reaches $68.8\%$ on VERINA and $62.3\%$ on AlgoVeri, with
implementations supplied by another model. The agent systems, which revise their output
on Lean's feedback, all prove $84.7\%$ or more of VERINA:
Aristotle, Aleph Prover, EasyProver and \eureka{}. \eureka{} reaches this with an open-weight model, and on AlgoVeri it
writes the implementation as well as the proof.

\begin{table}[!htbp]
\centering
\caption{\textbf{Reported results in program verification (\% proved).} Baselines from
Goedel-Code-Prover (its Fig.~3, pass@128) and NTP4VC (its Table~5, Lean,
pass@8 unless marked). Models, toolchains, supplied implementations and
budgets differ. Green shading scales with the share proved.}
\label{tab:code}
\footnotesize
\setlength{\tabcolsep}{5pt}
\setlength{\aboverulesep}{0pt}\setlength{\belowrulesep}{0pt}\setlength{\extrarowheight}{0.3ex}
\begin{tabular}{lccc}
\toprule
System & VERINA\,$\uparrow$\budget{(189)} & AlgoVeri\,$\uparrow$\budget{(77)} & NTP4VC\,$\uparrow$\budget{(600)} \\
\midrule
\multicolumn{4}{l}{\emph{General-purpose models}} \\
Claude-Opus-4.6         & \heat{4}{10.6} & \heat{5}{11.7} & \na \\
Gemini-3-Flash          & \heat{8}{19.6} & \heat{5}{11.7} & \na \\
GPT-5.2-Pro             & \heat{7}{16.9} & \heat{7}{18.2} & \na \\
GPT-5.3-Codex           & \heat{6}{15.9} & \heat{6}{14.3} & \na \\
DeepSeek-V3.2-Speciale  & \heat{5}{12.2} & \heat{2}{5.2} & \na \\
GPT-o4-mini-high        & \na  & \na  & \heat{0}{0.50\budget{@1}}  \\
DeepSeek-V3.1 / Qwen3-32B / Qwen3-235B-A22B & \na & \na & 1.67 / 1.17 / 1.00 \\
K2-think                & \na  & \na  & \heat{0}{0.00}  \\
\midrule
\multicolumn{4}{l}{\emph{Neural provers}} \\
Kimina-Prover-72B       & \heat{6}{15.9} & \heat{8}{19.5} & \na \\
DeepSeek-Prover-V2-671B & \heat{7}{18.5} & \heat{9}{23.4} & \heat{1}{3.00}  \\
Goedel-Prover-V2-32B    & \heat{7}{18.0} & \heat{9}{22.1} & \heat{1}{2.17}  \\
BFS-Prover-V2-32B       & \heat{10}{24.3} & \heat{8}{20.8} & \na \\
Goedel-Code-Prover-8B~\citep{goedelcodeprover2026} & \heat{28}{68.8\budget{@32}} & \heat{25}{62.3} & \na \\
\midrule
\multicolumn{4}{l}{\emph{Agent systems}} \\
Aristotle~\citep{harmonicverina2025}       & \heat{34}{84.7 (96.8$^{\dagger}$)} & \na & \na \\
Aleph Prover~\citep{alephverina2026}       & \heat{40}{100.0$^{\ddagger}$} & \na & \na \\
EasyProver~\citep{ganguly2026easyprover}   & \heat{40}{100.0$^{\dagger}$} & \heat{40}{100.0} & \heat{39}{98.0 (99.5$^{\dagger}$)}  \\
\midrule
\ours \eureka{} (ours)                     & \heat{40}{99.5 (100.0$^{\dagger}$)} & \heat{40}{\best{100.0}$^{\P}$} & \heat{39}{97.5 (99.0$^{\dagger}$)}  \\
\bottomrule
\end{tabular}
\par\vspace{1pt}
\parbox{\linewidth}{\scriptsize $^{\dagger}$Proofs plus refutations of the specification. $^{\ddagger}$Proof/refutation split not reported. $^{\P}$\eureka{} also writes the implementation; Goedel-Code-Prover receives one generated by another model.}
\end{table}

\subsection{Restarts by benchmark}
\begin{figure}[!htbp]
\centering
\includegraphics[width=\linewidth]{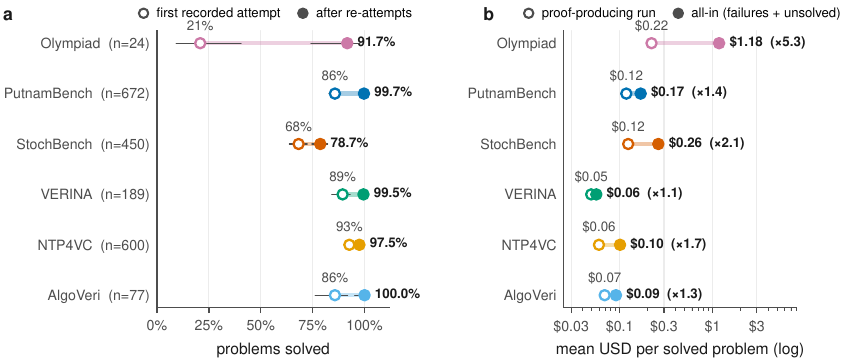}
\caption{\textbf{Coverage and cost before and after restarts.} (a)~Coverage after the first model-calling
attempt (open) and after continued runs (filled), with Wilson $95\%$
intervals. (b)~Mean cost per solved entry for the proof run alone (open) and
all-in (filled), including failed attempts and unsolved entries.}
\label{fig:retries}
\end{figure}

\Cref{fig:retries} breaks the effect of restarts down by benchmark. The first
model-calling run proves $1{,}679$ entries; $158$ more are proved on the second,
$33$ on the third, and $26$ after four to eight runs. Of the $217$ entries that
needed a restart, $73\%$ were proved by the first one. AlgoVeri goes from $85.7\%$ to $100\%$
and VERINA from $89.4\%$ to $99.5\%$, at a mean all-in cost $1.3$ and $1.1$ times
that of the proof run alone, so their failed first runs were cheap to
repeat. The olympiad sets gain the most, from $20.8\%$ to
$91.7\%$, and pay the most, $5.3$ times. StochBench gains $10.5$ points (from
$68.2\%$ to $78.7\%$) at $2.1$ times the cost, the highest cost ratio outside
the olympiad sets, consistent with the
definition gap described in \cref{sec:results}.

\subsection{The proof run alone}
\begin{figure}[!htbp]
\centering
\includegraphics[width=0.9\linewidth]{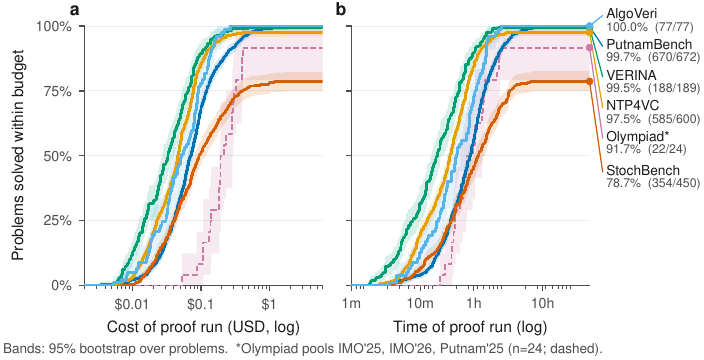}
\caption{\textbf{Coverage within a budget, proof run only.} As
\cref{fig:scaling}, but counting only the cost and time of the run that
produced the accepted proof, with the same denominator. Earlier unsuccessful
runs are omitted, so these curves describe the successful run;
\cref{fig:scaling} gives the full cost.}
\label{fig:scaling-success}
\end{figure}

\Cref{fig:scaling-success} repeats \cref{fig:scaling} counting only the run that
produced the accepted proof. The difference between the two is the cost of
the failed runs. For routine benchmarks the curves
nearly coincide: $80\%$ coverage costs $\$0.07$ on VERINA and $\$0.08$ on NTP4VC
for the proof run, against $\$0.08$--$\$0.09$ all-in, and on PutnamBench $\$0.14$
against $\$0.18$. On the olympiad sets the difference is large: the proof run
reaches $80\%$ at $\$0.35$, but $\$1.02$ all-in. A successful olympiad search costs
about two and a half times a successful PutnamBench one, and the attempts
before it nearly triple the cost of reaching $80\%$ coverage.

\subsection{Cost and time distributions}
\begin{figure}[!htbp]
\centering
\includegraphics[width=\linewidth]{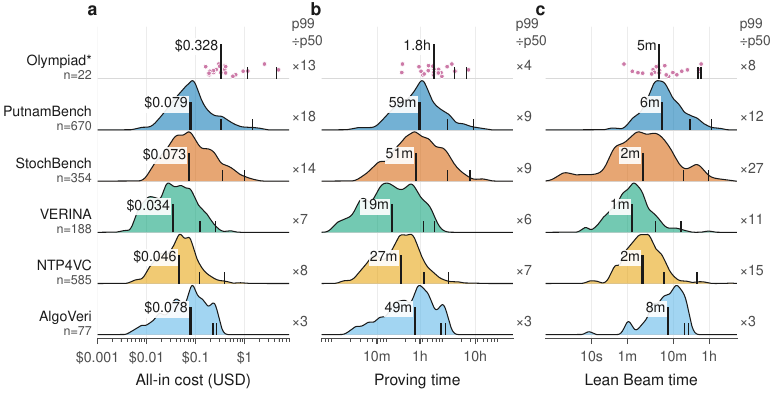}
\caption{\textbf{Distributions of cost, time and Lean time over solved entries.} Distributions over solved entries of (a)~all-in
cost, (b)~proving time and (c)~cumulative \beam{} time, as density estimates of
log values scaled to equal peaks. Tall ticks mark medians, short ticks the 90th
and 99th percentiles; the right column gives the 99th-percentile/median
ratio. The $22$ solved olympiad entries are shown as points.}
\label{fig:atlas}
\end{figure}

\Cref{fig:atlas} shows how cost, proving time and Lean time are distributed over
solved entries in each benchmark, and the ratio between the 99th percentile
and the median measures how heavy each tail is. Tail ratios are lowest on
program verification. AlgoVeri has the tightest spread in all three measures
(ratios $3.4$, $3.5$ and $2.8$), suggesting that writing and verifying a
specified program takes a fairly uniform amount of work, and VERINA and NTP4VC
follow with cost ratios of $7.4$ and $8.2$. Among the larger benchmarks,
PutnamBench has the highest cost ratio, $18.1$, and its median solve takes $59$ minutes against $9.3$ hours at the 99th percentile. StochBench's
Lean time is the most variable of all (ratio $26.7$): most of its proofs need
only about two minutes of Lean checking, while the hardest need close to an
hour. AlgoVeri's median solve spends $7.6$ minutes in \beam{} calls out of $49$, more than any
other benchmark in relative terms, likely because the agent must also check
the behavior of the program it writes. Capacity planning for a public service
must cover these tails: the 99th-percentile PutnamBench solve runs $9.3$ hours.

\FloatBarrier
\subsection{Accepted proofs}
\begin{figure}[!h]
\centering
\includegraphics[width=0.95\linewidth]{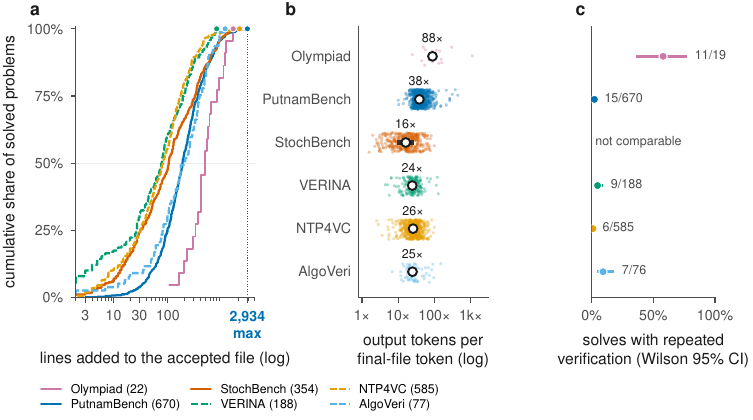}
\caption{\textbf{Size of accepted proofs.} (a)~Cumulative distribution of lines added to the accepted file
relative to the original statement; dots mark each group's longest proof.
(b)~Output tokens generated per token of the final file (four characters per
token). (c)~Share of solved entries, among those with recorded checks, that
went through more than one verification before acceptance (Wilson $95\%$
intervals). StochBench logs checks differently and is not shown.}
\label{fig:proofs}
\end{figure}

Accepted files add a median of $125$ lines to the original statement, $536$ at
the 90th percentile and up to $2{,}934$ (\cref{fig:proofs}). Olympiad proofs are the longest, with a median of $479$ added lines,
followed by AlgoVeri ($206$), whose files contain both an implementation and
its proof, and PutnamBench ($187$). VERINA and NTP4VC proofs are shorter
($77$ and $81$ lines), since the program is supplied and the obligations are
local. The ratio of generated output to final file indicates how much output
was generated per token kept. It is lowest on StochBench ($16$), perhaps because
hypotheses supply much of the structure, around $25$ on program verification, $38$ on
PutnamBench and $88$ on the olympiad sets: on olympiad problems the model
generated about $88$ output tokens for each token of the final file. $11$ of
the $19$ solved olympiad entries with recorded checks went through
more than one verification before acceptance.

\end{document}